\documentclass[12pt]{iopart}

\usepackage[latin1]{inputenc}
\usepackage[dvips]{graphicx}
\usepackage{bm}
\usepackage{hyperref}
\usepackage{ulem}

\usepackage{color}
\definecolor{greencolor}{rgb}{0,0.5,0.2}
\definecolor{redcolor}{rgb}{0.,0.,0.}
\definecolor{bluecolor}{rgb}{0,0.,1.}
\definecolor{greycolor}{rgb}{.5,.5,.5}
\def\Red#1{{\color{redcolor} #1}}

\begin{document}

\title{Structure-Semantics Interplay in Complex Networks and Its Effects on the Predictability of Similarity in Texts}

\author{Diego Raphael Amancio$^1$, Osvaldo N. Oliveira Jr.$^1$, Luciano da Fontoura Costa$^1$}

\address{$^1$  Institute of Physics of S\~ao Carlos \\
	University of S\~ao Paulo, P. O. Box 369, Postal Code 13560-970 \\
	S\~ao Carlos, S\~ao Paulo, Brazil \\
}


\ead{diego.amancio@usp.br,diegoraphael@gmail.com}

\begin{abstract}
The classification of texts has become a major endeavor with so much electronic material available, for it is an essential task in several applications, including search engines and information retrieval. There are different ways to define similarity for grouping similar texts into clusters, as the concept of similarity may depend on the purpose of the task. For instance, in topic extraction similar texts mean those within the same semantic field, whereas in author recognition stylistic features should be considered. In this study, we introduce ways to classify texts employing concepts of complex networks, which may be able to capture syntactic, semantic and even pragmatic features. The interplay between the various metrics of the complex networks is analyzed with three applications, namely identification of machine translation (MT) systems, evaluation of quality of machine translated texts and authorship recognition. We shall show that topological features of the networks representing texts can enhance the ability to identify MT systems in particular cases.
For evaluating the quality of MT texts, on the other hand, high correlation was obtained with methods capable of capturing the semantics. This was expected because the golden standards used are themselves based on word co-occurrence. Notwithstanding, the Katz similarity, which involves semantic and structure in the comparison of texts, achieved the highest correlation with the NIST measurement, indicating that in some cases the combination of both approaches can improve the ability to quantify quality in MT. In authorship recognition, again the topological features were relevant in some contexts, though for the books and authors analyzed good results were obtained with semantic features as well. Because hybrid approaches encompassing semantic and topological features have not been extensively used, we believe that the methodology proposed here may be useful to enhance text classification considerably, as it combines well-established strategies.
\end{abstract}

\newpage
\tableofcontents

\maketitle

\section{Introduction}
\label{}

The growing amount of text electronically available has placed Natural Language Processing (NLP) in the spotlight~\cite{science,spot2,spot3}. Many are the applications exploiting NLP, including machine translation~\cite{mach1}, automatic summarization~\cite{sum3}, search engines~\cite{eng1,eng2}, writing tools~\cite{aluisio1}, text simplification~\cite{simpl1}, information retrieval~\cite{inf1}, in addition to various resources such as  thesaurus~\cite{wordnet} and corpus~\cite{corpus1}. In some of these applications, one needs to estimate the similarity between documents. Indeed, summarizers and translators are usually evaluated according to the similarity with a reference text produced by humans, while categorizers and clustering applications~\cite{kohonen} employ similarity measures to establish clusters containing similar texts. Defining similarity is not straightforward, though.

Due to the practical and even theoretical (since the computation of similarity involves understanding the cognitive processes) interests related to the estimation of pairwise similarities, a wide variety of indices have been developed. The vast majority are based on semantic similarity, which is calculated by counting the number of keywords or n-grams shared by two documents ~\cite{damashek}. More sophisticated techniques based on semantic analysis~\cite{semant2} have also been used which go beyond counting the number of shared words in distinct texts. Although such methods can be considered efficient because they have a reasonable correlation with the human assessment~\cite{empirical}, for some applications a semantic analysis may not suffice, since the textual structure plays a prominent role. In classifying different literary styles, for example, there may be a correlation between the theme of different styles, but textual structure is expected to be a key factor to characterize the styles~\cite{yang}. Therefore, similarity indices based on text structure may be more useful in this type of application. Analogously, the structure-based similarity indices could be useful for clustering texts with the same quality of writing~\cite{quality}; quality of translation~\cite{t1,t2} and even texts endorsing the same point of view~\cite{op1}, since all these applications can be suitably characterized by the structural paradigm.

%

Since both the style and the semantics can be useful for estimating similarity, in this article we study the interplay between semantics and structure in 3 NLP applications: (i) evaluation of quality of translations, (ii) translation classification (i.e., identification of which machine translator generated a given translation) and (iii) authorship recognition. Using formalisms based on the representation of texts as complex networks we derived (dis)similarity indices based on semantics and structure to show that both types of indices are able to reveal patterns that would be hidden if only one of the paradigms were used.

\section{Complex Networks and Natural Language Processing}

\Red{
Concepts and methods from complex networks have been employed to analyze many aspects of language~\cite{spot3,quality,t1,t2,extractive}, including analysis of syntactic networks~\cite{chinesedependence,patternscancho}, classification of languages through topological analysis~\cite{class1,class2,class3} and investigation of phonetic aspects~\cite{fonetico}. Even though the main aim in using complex networks has been to study linguistic phenomena, in the majority of the studies the semantics has been disregarded because the focus is normally on the topology of the network. On the other hand, typical applications of natural language processing~\cite{manning} only consider the semantic relationship between documents, as is the case of methods to quantify pairwise similarity, such as the bag-of-words technique~\cite{manning}. In the latter, the number of shared terms is used as an estimate of similarity, while the order in which words occur is neglected. Obviously, a more thorough analysis of linguistic phenomena should be expected if one is able to combine semantic with topological features.
}

\Red{
Indeed, hybrid approaches have proven promising not only for text analysis but also for modeling networks. For example, Menczer~\cite{menczer} showed that models of citation networks are more accurate if in addition to topological characteristics (e.g, the degree distribution) they include features regarding content similarity. Likewise, Mehler~\cite{mehler} showed that the interplay between semantic and structural features emerging from social networks is essential for representing and classifying large complex networks. Along these lines, in this paper we propose methods to estimate text similarity taking into account both \emph{topological and semantic} features. With topology one may capture stylistic features concerning authorship~\cite{uzuner}, complexity~\cite{complexidade}, quality~\cite{quality} and aspects that depend on non-trivial relationships between textual concepts. As for the semantic investigation, our aim is to capture textual pragmatic features so that manuscripts are clustered together when they share a given topic. As we shall show, the combined use of both strategies leads to improved evaluation of machine translation systems because the vast majority of established quality indexes neglect long-range stylistic information.
}

\section{Methodology} \label{matmet}

\subsection{Modeling Texts as Networks} \label{modeling}

The model used in this work was adopted in many previous studies~\cite{quality,t1,t2,op1}. The process starts by eliminating words conveying low semantic content (the full list of stopwords is given in the Supplementary Information (SI)) for we are interested only in the relationship between content words. \Red{It is true that that by removing stopwords one may miss out on very important linguistic information~\cite{stop1,stop2}. However, in the approaches we used, removing stopwords is entirely justified for two reasons: 1) The statistics of the metrics in the complex networks would be unduly affected by the highly frequent stopwords (such as articles and prepositions) that may be connected with any type of node. 2) The stopwords were removed to make the topological analysis consistent with the semantic analysis, where stopwords play no role on the prediction of pairwise similarities ~\cite{manning}.} The remaining words are transformed to their canonical form, where verbs are converted to their infinitive form and nouns are converted to the singular form. After this step, each distinct word becomes a node of the network. Edges link two words if they appear as neighbors in the pre-processed text. Section $2$ of the SI illustrates step-by-step the construction of the network. Mathematically, the network is represented by a weighted matrix $W$, where $w_{ij}$ stores the number of times the word $i$ appeared before the word $j$. Alternatively, we also use the unweighed and undirected representation of $W$, which is known as adjacency matrix $A$. If words $i$ and $j$ are neighbors in the text, then $a_{ij} = 1$. Otherwise, $a_{ij} = 0$ otherwise.

\subsection{Topological Network Measurements} \label{medidas}

The topological measurements described in this section are associated with topological features of the network, with no concern for the semantics of the nodes. These metrics are degree $k$, betweeness $B$, the average shortest path length $l$ and the clustering coefficient $C$. All these measurements have been widely employed in the topological characterization of complex networks~\cite{costa2}. They will be employed in the derivation of the so called topological similarity indices in Section \ref{topIndexes}. More details about these measurements are given in Ref.~\cite{costa2} and in Section 2 of the SI.


\subsection{Topological Network Motifs}

The network structure can also be characterized using motifs~\cite{motif2}, which can be taken as small building blocks comprising nodes and edges.  They may be seen as subgraphs that appear on the network more than expected just by chance. To verify if the frequency of a particular motif $m$ is higher than expected, the z-score measurement $Z$ is used. To compute $Z$, let $r_{m}$ and $\sigma_{m}$ be, respectively, the average and the standard deviation of the frequency of $m$ over 100 random equivalent networks (i.e., the random networks have the same number of vertices and edges as the original network). If $n_m$ is the frequency of $m$ in the network of interest, then $Z$ is given by:

\begin{equation}
 Z(m) = \frac{n_m - r_m}{\sigma_m}
\end{equation}

\begin{figure}[h]
		\begin{center}
			\includegraphics[width=0.7\textwidth]{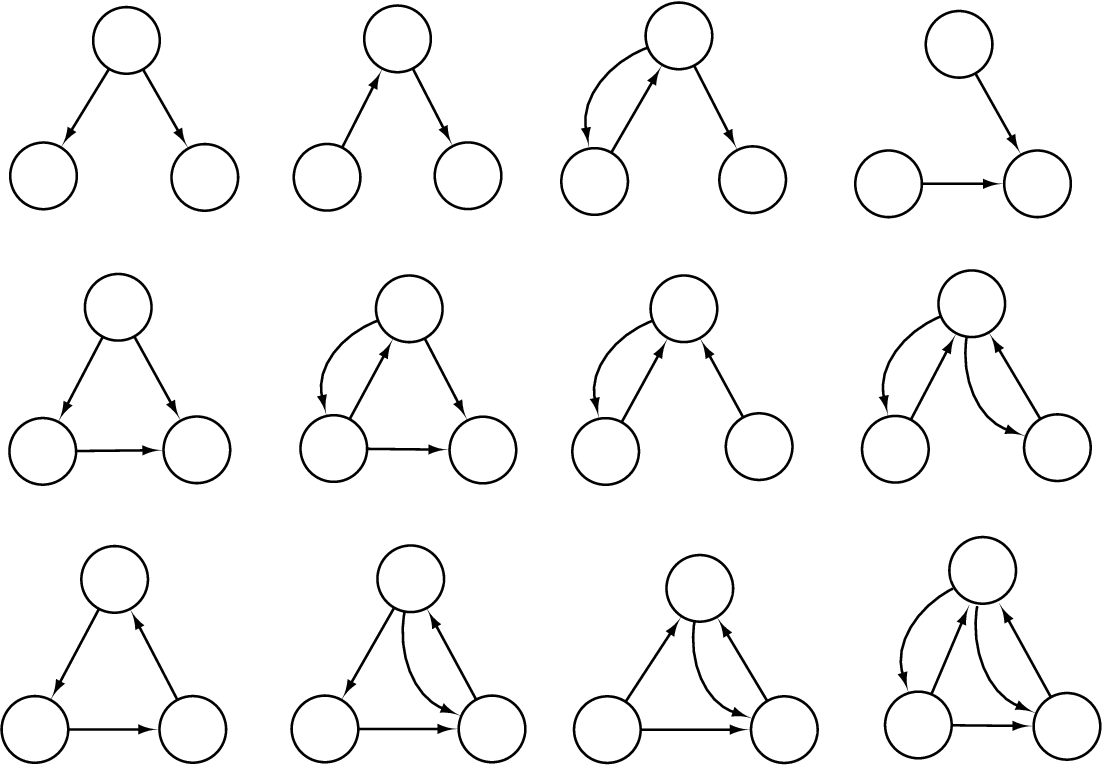}
		\end{center}
		\caption[\it]{\it Motifs employed to characterize the topology of the networks through the analysis of z-scores.}
		\label{fig.b2}
\end{figure}

It is known that some families of networks display high frequency of special motifs. Therefore, one may identify the function of the network by examining the frequency or significance profile of each motif~\cite{newmanbook}. In this work, we focused on motifs involving three vertices, as shown in figure \ref{fig.b2}. However, any other motif of particular interest could have been used.

\section{Similarity and Dissimilarity Indices}

In this section we derive the indices used to evaluate and classify machine translations and recognise authorship of prose and poetry. The indices can be divided into three groups, according to the use of topological or semantics features. If no topological measurements are used to define the index, then it is solely based on semantics. On the other hand, if no information on the label of the vertices is used, then the index is entirely topological. Finally, if both topological and semantic features are employed, the index is considered as a hybrid.

\subsection{Topological Dissimilarity Indices}

\subsubsection{Dissimilarity Index Based on Topological Network Features}
\label{topIndexes}

This index is computed by obtaining the distance $d_{st}$ of a given text $T_s$ to a reference text $T_t$. Let $\overrightarrow{\mu}$ be a vector where each component represents the average for one measurement described in Section \ref{medidas}. The vectors $\overrightarrow{\mu_s}$ and $\overrightarrow{\mu_t}$, related to $T_s$ and $T_t$, respectively, are compared, leading to the vector $\overrightarrow{\delta}$. The difference is computed component by component:

\begin{equation} \label{diff1}
	\delta(i) = \frac{|| \mu_t(i) - \mu_s(i) ||}{\mu_s(i)}
\end{equation}

	The distance $d_{st}$ is then obtained as the average of the differences:

\begin{equation}
	d_{st} = \frac{1}{n} \sum_{i=1}^{n} \delta(i),
\end{equation}
where each component $i$ of $\overrightarrow{\delta}$ is the difference for a given measurement. Then, each component can be considered as a dissimilarity index itself.

\subsubsection{Dissimilarity Index based on Motifs}

Similarly to the previous one, this index is defined as the average over the differences in $\overrightarrow{\delta}$, with each component $\delta(i)$ being the z-score obtained for the motif associated with index $i$:

\begin{equation}
	\delta(i) = \frac{|| Z_t(i) - Z_s(i) ||}{Z_s(i)}
\end{equation}

\subsection{Semantic-Based Indices} \label{semanticIndexes}

	In this section we describe the indices based solely on the semantics. In other words, only the label and the information of the immediate neighborhood of the nodes are taken to quantify pairwise similarities. As in previous cases, all indices in this section consider that two nodes are similar if they share many neighbors.

   When labels are used to compare networks of texts according to the number of shared neighbors, a problem arises if a node with a specific label does not appear in the other network. To obviate this problem, we adopted a strategy where a minimum similarity value is assigned to nodes that appear in only one of the networks. This can be obtained in the following manner. Let $A^s$ and $A^t$ be the networks being compared. If a given node belonging to $A^s$ has label $L$ and if $A^t$ has no node with label $L$ then a new node is created in $A^ t$ without any connections. Thus, since the node labeled with $L$ in $A^t$ has no neighbors, it will have no shared neighbors. Consequently, the similarity related to this node will be zero.

\subsubsection{Cosine Similarity}

The number of sharing neighbors $q_{ii}$ of two nodes $v^s_i$ and $v^t_i$ with the same label $L_i$ in the networks $A^s$ and $A^t$ (which refer to $T_s$ and $T_t$, respectively) is:

\begin{equation}
	q_{ii} = \sum_{j}  A^s_{ij} A^t_{ij}.
\end{equation}


	While this measure captures the number of common neighbors, it is difficult to apply because one does not know whether $q_{ii}$ is small or large~\cite{newmanbook}. In fact, a better approach should first normalize $q_{ii}$ to confine its range within a strict interval. To perform such normalization, we divide $q_{ii}$ by the geometrical mean between the degrees $k_i^s$ and $k_i^t$ of the nodes considered:
\begin{equation} \label{cosine}
	c_{ii} = \frac{\sum_{k} A^s_{ik} A^t_{ik}}{\sqrt{\sum_{k} A^s_{ik}}\sqrt{\sum_{k} A^t_{ik}} } = \frac{q_{ii}}{\sqrt{k^s_i k^t_i}}
\end{equation}
Note that $c_{ii}$ can be interpreted as the cosine of the angle between the vectors of neighbors. Therefore, the higher $c_{ii}$ the smaller the angle and consequently the larger the similarity is.

\subsubsection{Similarity Index Based on the Pearson Correlation Coefficient}

The cosine similarity above is an effective way to normalize $q_{ii}$ in the sense that it limits its range in the interval between zero and 1, but other normalizations can be considered. For example, Ref.~\cite{newmanbook} suggests that $q_{ii}$ should be compared to the expected number $\sigma_{ii}$ of sharing neighbors, supposing that the choice of neighbors are made randomly. To quantify this expected value, let $k^s_i$ and $k^t_i$ be the degree of $v^s_i$ and $v^t_i$ being compared, respectively. If $v^s_i$ randomly chooses its neighbors, the probability that it chooses a node that is also neighbor of $v^t_i$ is equal to $k^t_i/n$. Repeating the process for the remaining $k_i - 1$ neighbors of $v^s_i$, the expected number of sharing neighbors will be $\sigma_{ii} = k^s_i k^t_i/n$. Thus, the similarity $\Sigma_{ii}$  can be computed as the difference between the actual number of shared neighbors $q_{ii}$ and $\sigma_{ii}$, which leads to:

\begin{eqnarray} \label{pearson}
	\Sigma_{ii} &=& q_{ii} - \sigma_{ii} = \sum_{k}  A^s_{ik} A^t_{ik} - \frac{k^s_i k^t_i}{n} \nonumber \\
			&=& \sum_{k}  A^s_{ik} A^t_{ik} - n  \overline{k}^s_i  \overline{k}^t_i \nonumber \\
			&=& \sum_{k} \Big{(}  A^s_{ik} A^t_{ik} - \overline{k}^s_i  \overline{k}^t_i \Big{)} \nonumber \\
			&=& \sum_{k} \Big{(} A^s_{ik} -  \overline{k}^s_i \Big{)} \Big{(} A^t_{ik} -  \overline{k}^t_i \Big{)},
\end{eqnarray}
where the notation $\overline{k}$ represents the degree normalized by the number of nodes in the network:
\begin{equation}
	\overline{k}_i = \frac{1}{n} \sum_{k} A_{ik}.
\end{equation}
Eq. (\ref{pearson}) can be seen as a covariance, i.e., a non-normalized correlation, which can be normalized by dividing by standard deviations of the vectors $A^s_{ik}$ and $A^t_{ik}$, $k$ = 1 .. $n$. Performing such normalization, the covariance becomes the Pearson correlation measure $\rho_{ii}$:

\begin{equation}
	\rho_{ii} = \frac{\sum_{k} \Big{(} A^s_{ik} -  \overline{k}^s_i \Big{)} \Big{(} A^t_{ik} -  \overline{k}^t_i \Big{)}}{ \sqrt{\sum_{k} \Big{(} A^r_{ik} -  \overline{k}^s_i \Big{)}^2} \sqrt{\sum_{k} \Big{(} A^t_{ik} -  \overline{k}^t_i \Big{)}^2} },
\end{equation}
which ranges between -1 and 1. Just to keep the range of the similarity metrics in the interval between zero and 1, the following linear transformation was performed in $\rho_{ii}$, deriving $\rho_{ii}'$:
\begin{equation}
	\rho_{ii}' = \frac{\rho_{ii} + 1}{2}
\end{equation}
Thus to interpret if $\rho_{ii}'$ is high, it is sufficient to verify $\rho_{ii}' > 0.5$, since the threshold 0.5 corresponds to the similarity obtained when the number of shared neighbors is the same as expected by chance.

\subsubsection{Leicht-Holme-Newman Index} 

An alternative to quantify how the number of shared neighbors is greater than expected would be to check the ratio between the actual and expected values, instead of calculating the difference, as was done in the derivation of $\Sigma$ in eq. (\ref{pearson}). In this case, the similarity coefficient, referred to as Leicht-Holme-Newman Index~\cite{prediction}, is given by:

\begin{equation}
	\tau_{ii} = \frac{q_{ii}}{\sigma_{ii}} = n \frac{\sum_{k} A^s_{ik} A^t_{ik}}{\sum_{k} A^s_{ik} \sum_{k} A^t_{ik}}
\end{equation}

The threshold to be analyzed is 1. If $\tau_{ii} $ is above 1, the similarity is higher than expected. Otherwise, the value is less than 1 but always positive. It is still worth noting the resemblance of $\tau$ with $c$ defined in eq. (\ref{cosine}), since while the former divides $q_{ii}$ by $k^s_i k^t_i$, the latter divides $q_{ii}$ by $\sqrt{k^s_i k^t_i}$. Even though in principle the difference between these measures is small, some authors suggest that $\tau$ is far more effective since it presents a well-defined threshold to interpret similarity~\cite{newmanbook,prediction}.

\subsubsection{Similarity Based on the Euclidean Distance} \label{euclideana}

This measure was derived using again the neighboring vectors $A^s_{ik}$ and $A^t_{ik}$. The similarity between the two vectors is obtained by calculating the euclidean squared distance between them. The distance is then normalized by the maximum possible distance $k^s_i+k^t_i$, which occurs when there are no shared neighbors. Thus the resulting distance can be expressed as:

\begin{equation} \label{euclideano}
	d_{ii} = \frac{\sum_{k} \Big{(} A^s_{ik} - A^t_{ik} \Big{)} ^ 2}{ k^s_i  +  k^t_i } = \frac{\sum_{k} \Big{(} A^s_{ik} + A^t_{ik} - 2 A^s_{ik}  A^t_{ik} \Big{)} }{k^s_i  +  k^t_i} = 1 - 2 \frac{q_{ii}}{k^s_i  +  k^t_i }
\end{equation}

	Note that once again the index ranges from 0 to 1. Therefore, to obtain the corresponding similarity index $\kappa_{ii}$, the complement is taken:

\begin{equation}
	\kappa_{ii} = 1 - d_{ii} = 2 \frac{q_{ii}}{k^s_i  +  k^t_i }.
\end{equation}

	Interestingly, $\kappa_{ii}$ can be seen as a variation of $c_{ii}$, since while the latter normalizes $q_{ii}$ by the geometric mean, the former normalizes $q_{ii}$ using the arithmetic mean.

\subsection{Similarity Indices Based on Both Topological and Semantic Features}

\subsubsection{Katz Similarity}

	While in Section \ref{semanticIndexes} the number of shared neighbors played a prominent role in quantifying the similarity, the index derived here is based on the idea that two vertices are similar if the neighbors of one node are similar to the neighbors of the other node. In other words, two nodes need not share the same neighbors to be considered similar, they only need to have neighbors which are similar. To develop the measure, the similarity between all possible pairs of nodes of the network is computed. Then, the correlation of such similarity is verified for the nodes with the same label (i.e., they refer to the same word) in the networks being compared. The description of these two steps is given below.

	Initially, to store the similarity between the pairs of nodes $i$ and $j$ of one of the two networks being compared, the variable $\varsigma_{ij}$ is created. Assuming that $\varsigma_{ij}$ is proportional to the similarity of the corresponding neighbors $k$ and $l$, then $\varsigma_{ij}$ can be recursively defined as:

\begin{equation} \label{aba}
	\varsigma_{ij} = \alpha \sum_{k}^{} \sum_{l}^{} A_{ik} A_{jl} \varsigma_{kl},
\end{equation}

Even though $\varsigma_{ij}$ seems a consistent measure,~\cite{newmanbook} highlights some drawbacks arising from this definition. For example, $\varsigma_{ij}$ not necessarily takes high values when the self-similarity $\varsigma_{ii}$ is computed. Consequently, nodes with many common neighbors can be overlooked. To solve this problem,~\cite{newmanbook} suggests adding an artificial term to ensure that $\varsigma_{ii}$ takes high values. The modification leads to a new definition of $\varsigma$:

\begin{equation} \label{original}
		\varsigma_{ij} = \alpha \sum_{k}^{} \sum_{l}^{} A_{ik} A_{jl} \varsigma_{kl}  + \delta_{ij}
\end{equation}

Isolating $\varsigma$ in eq. (\ref{original}), it can be written as a summation of the number of paths of even length connecting the nodes $i$ and $j$. Obviously, there is no reason for using only paths of even lengths. For this reason, $\varsigma$ is redefined as:
\begin{equation}
		\varsigma_{ij} = \alpha \sum_{k}^{} \sum_{l}^{} A_{ik}  \varsigma_{kj}  + \delta_{ij},
\end{equation}
which leads to the following closed form:
\begin{equation} \label{sooma}
	\varsigma = ( I - \alpha A ) ^ {-1} = \sum_{i=0}^{\infty} (\alpha A) ^ i,
\end{equation}
where $I$ represents the identity matrix. Assuming that $i$ and $j$ are similar if $i$ has $k$ as neighbor and $k$ is similar to $j$, then the closed solution also takes into account paths of odd lengths, since the summation is performed over the integers. With regard to the $\alpha$ parameter, to guarantee that the summation in eq. (\ref{sooma}) converges, it must lie in the interval $\alpha < \lambda_1^{-1}$, where $\lambda_1$ is the largest eigenvalue of A. In particular, we have chosen $\alpha = \lambda_1^{-1} / 2$.

After calculating the similarity between all pairs of nodes of the networks under comparison, we compute the similarity using the Hubert's coefficient:
\begin{equation} \label{hubert}
	\Gamma = \frac{1}{\Delta_s \Delta_t} \frac{2}{N(N-1)} \sum_{i=1}^{N-1} \sum_{j=i+1}^{N} (\varsigma^s_{ij} - \mu_s)(\varsigma^t_{ij} - \mu_t),
\end{equation}
where $\mu$ and $\Delta$ are given by:
\begin{equation}
	\mu =  \frac{2}{N(N-1)} \sum_{i=1}^{N-1} \sum_{j=i+1}^{N} \varsigma_{ij}
\end{equation}
\begin{equation}
	\Delta =  \frac{2}{N(N-1)} \sum_{i=1}^{N-1} \sum_{j=i+1}^{N} ( \varsigma_{ij} - \mu )^2
\end{equation}

Upon defining $\Gamma$ as shown in eq. (\ref{hubert}), the networks are considered similar to each other in case the numbers of paths between every pair of nodes are correlated. In other words, $\Gamma$ is high when strongly connected pairs of nodes in one of the networks tend to be also strongly connected in the other network. Analogously, $\Gamma$ also takes high values when weakly connected pairs of nodes in the first network are weakly connected in the second network.

\subsubsection{Similarity Based on the Ability to Match Nodes}

One of the recent areas of research in complex networks encompasses the analysis of the interrelationship between networks, which contrasts with the study of isolated networks. For example, in communication networks, there exists a duality between online acquaintanceship networks and the network of phone contacts. In fact, this happens precisely because both networks are actually social networks~\cite{social1,social2,social3,social4}. In language networks, the same effect occurs, since a given word can display the same pattern in different languages, especially if the languages have a common origin~\cite{t1}. Based on this interrelationship, we developed a similarity index which works in two steps. First, a heuristic is applied to perform the matching between nodes of the networks. Then, the quality of matching is evaluated by counting the number of accurate matching (i.e. the number of associations in which the associated labels correspond to the same word). In particular, we assume that the similarity is directly proportional to this accuracy rate, since similar texts are expected to share semantic as well as topologic properties.

	The method employed to map nodes previously discussed can be separated into two steps:

\begin{list}{}{}
\item (i) \emph{Computation of similarity}: the similarity between two vertices of distinct networks can be computed using structural or semantic information. In the first case, referred to as topologic matching, the similarity is calculated using the local relative difference of topologic measurements, as defined in eq. (\ref{diff1}). In the second case, referred to as semantic matching, we used the cosine similarity defined in eq. (\ref{cosine}).

\item 	(ii)  \emph{Mapping}: representing similarities computed in (i) as a bipartite network where the weights of the links represent the similarities, we applied the KM algorithm~\cite{hungarian} in order to find the pairs which maximize the sum of the matching links. Actually, this algorithm does not always find the best matching, since it is a heuristic to avoid evaluating all possibilities and enhance the efficiency in processing time. Nevertheless, more often than not the matching found by the heuristic is very similar to the best matching~\cite{combina}.
\end{list}

\subsubsection{Similarity Indices Based on the Preservation of Local Measurements (Slope) }

Similarly to the previous measure, we used a mapping to evaluate the similarity between texts. However, while the previous one assesses the accuracy rate of the mapping from the information of the similarity between all pairs of nodes, this one evaluates the variation of the topological measurements knowing in advance the correct mapping. That is to say, the measurements for words with the same labels in the networks are compared. The procedure to perform this comparison begins by plotting the measurements extracted from both texts, node by node, so each distinct measurement leads to a scatter plot. Thus if a given measurement $\mu$ is computed for node $v$, $\mu^{t}_v$ represents the value of $\mu$ for $v$ in one network and if $\mu^{s}_v$ represents $\mu$ for the same node in the other network, then the point ($\mu^{t}_v$,$\mu^{s}_v$) will belong to the scatter plot. Three descriptors are extracted from each scatter plot: the y-intercept, the slope and the Pearson product-moment correlation coefficient, obtained from the best straight line approximated by the least squares method. These descriptors are important because information about the preservation of metrics can be obtained.  In particular, if the y-intercept is close to zero and the slope and Pearson are close to 1, then the texts are similar to each other. Although they cannot be considered as self-contained similarity indices (since they are mutually dependent), these indices are still useful to capture the ability to preserve local measurements. In fact, to illustrate the use of such measures and compare them with the other similarity indices, we computed the three coefficients for each of the topological measurement and used them as attributes of the texts.

\section{Results and Discussion} \label{resultados}

\subsection{Evaluation of the Quality of Machine Translation} \label{ev_quality}

Evaluating the quality of MT has been as important as it is a difficult task. For obvious reasons, human evaluation is the most reliable, but it is too costly for large scale use, in addition to the problems of lack of agreement among distinct evaluators. These difficulties have motivated the development of several parameters to assess the quality of MT systems, the most used of which have been BLEU~\cite{bleuref} (Bilingual Evaluation Understudy) and NIST~\cite{nistref} (National Institute of Standards and Technology) indices. The latter quantify the quality of a translation according to the number of words appearing in both the translation and in one or more reference texts. Significantly, these parameters have been shown to correlate well with human judgment~\cite{bleuref}.

In our analysis we shall assume that high BLEU or NIST scores mean high quality of the translation, and therefore these scores will be a sort of golden standards. To verify the ability of the indices proposed to quantify similarity, we calculated the Pearson correlation between these indices and the two golden standards. The closer to 1 the Pearson coefficient the more appropriate is the parameter to quantify the translation quality.

In the experiments, we used a set of $100$ pieces of text compiled manually from the online edition of the Brazilian magazine Pesquisa FAPESP~\cite{fapesp}. The magazine is also available in Spanish~\cite{fap_esp} and English~\cite{fap_eng}, and therefore these human translations were used as reference (referred to here as golden standards). The translations to be evaluated were generated with the following machine translators: Google~\cite{googletranslator} (for Portuguese - English and Portuguese - Spanish), Bing~\cite{bingtranslator} (Portuguese - English and Portuguese - Spanish), Apertium~\cite{apertiumtranslator} (Portuguese - Spanish) and InterTran~\cite{intertrantranslator} (Portuguese - English). Thus, for each pair of languages, we obtained $300$ translations (target texts), $100$ original texts (source texts) and $100$ reference texts (golden standards).

The results for the pairs Spanish - Portuguese and English - Portuguese are shown in Tables \ref{tab.0} and \ref{tab.1}, respectively. The indices based solely on co-occurrence of words, which basically amounts to capturing semantic features, give the highest correlations (above 0.9). This result was expected because the BLEU and NIST scores are themselves based on co-occurrence of words between the translated text and a reference text. With regard to the indices using topological measurements, the Katz parameter was the only one to provide high correlation with BLEU and NIST scores. It seems that somehow the number of paths between concepts is useful to evaluate the quality of translations in the same way that it has been useful to characterize the topology of various networks~\cite{frank}. To further explore this ability to predict the semantic quality through structural analysis, we also calculated the correlation between the variation of each global measurement of the complex networks and the scores BLEU and NIST (results not shown). In both language pairs, we observed reasonable correlations for the variation of the standard deviation of betweenness (about 0.55). For the pair Spanish - Portuguese, we also observed a correlation between the standard deviation of out degree (about 0.52). This means that even without making use of any information about the nodes labels, it is still possible to predict with reasonable accuracy the semantic quality of translations, which depends on the information on labels.

\begin{table}
	\centering
	\caption{\label{tab.0} Absolute values for the correlation between the (dis)similarity indices based on complex networks and the golden standards BLEU and NIST for the translations from Spanish into Portuguese.  Semantic, Topologic and Hybrid measures are identified as (S), (T) and (S+T), respectively. As expected, the predominantly semantic metrics correlate better with the golden standards. Interestingly, the Katz metric, which uses topological measurements to compare texts, also correlates strongly with BLEU and NIST. }
		\begin{tabular}{|c|c|c|}
			\hline
			\textbf{(Dis)similarity Index}	&\textbf{BLEU}		&\textbf{NIST}  \\
			\hline
			Semantic Matching (S)		& 0.95 		& 0.61		\\
			Cosine				(S)	   & 0.94 		& 0.70		\\
			Pearson				(S)	   & 0.93 		& 0.68		\\
			Euclidean			(S)	   & 0.91 	   & 0.66		\\
			Katz					(S+T)	   & 0.81 		& 0.78		\\
			Leicht-Holme-Newman (S)	   & 0.70 		& 0.44		\\
			Topologic Matching  (S+T)   & 0.61      & 0.26		\\
			Motifs					(T)   & 0.36		& 0.26		\\
			\hline
		\end{tabular}
\end{table}

\begin{table}
	\centering
	\caption{\label{tab.1} Absolute values of the correlation between the (dis)similarity indices based on complex networks modeling and the golden standards BLEU and NIST for the translations from Portuguese into English. Semantic, Topologic and Hybrid measures are identified as (S), (T) and (S+T), respectively. As expected, the predominantly semantic metrics correlate better with the golden standards. Similarly to the result obtained for Spanish and English, the Katz similarity index also correlates strongly with BLEU and NIST.}
		\begin{tabular}{|c|c|c|}
			\hline
			\textbf{(Dis)similarity Index}	&\textbf{BLEU}		&\textbf{NIST}  \\
			\hline
			Cosine (S)					&	0.97 		&	0.79	\\
			Pearson (S)					&	0.96 		&	0.78	\\
			Euclidean (S)				&	0.95 		&	0.76	\\
			Katz	(S+T)					&	0.94		& 	0.87	\\
			Semantic Matching (S)		&	0.94		& 	0.80  \\
			Leicht-Holme-Newman (S)	&	0.54 		& 	0.38	\\
			Topologic Matching (S+T)	& 	0.37 		& 	0.20	\\
			Motifs				(T)	&	0.33		& 	0.18	\\
			\hline
		\end{tabular}
\end{table}

One may therefore conclude that topology can be useful to quantify similarity in this type of application. Of special relevance is the Katz index, which besides being the only metric using topology that achieved a reasonably high correlation with the BLEU metric, it was also the index that best correlated with the NIST index (see Table 3). Thus, the hypothesis that stylistic factors combined with semantic factors may be useful for evaluating the quality of automatic translations (especially for the Katz index to predict NIST) is confirmed.

\subsection{Classifying Translations}

\begin{figure}[h]
		\begin{center}
			\includegraphics[width=0.85\textwidth]{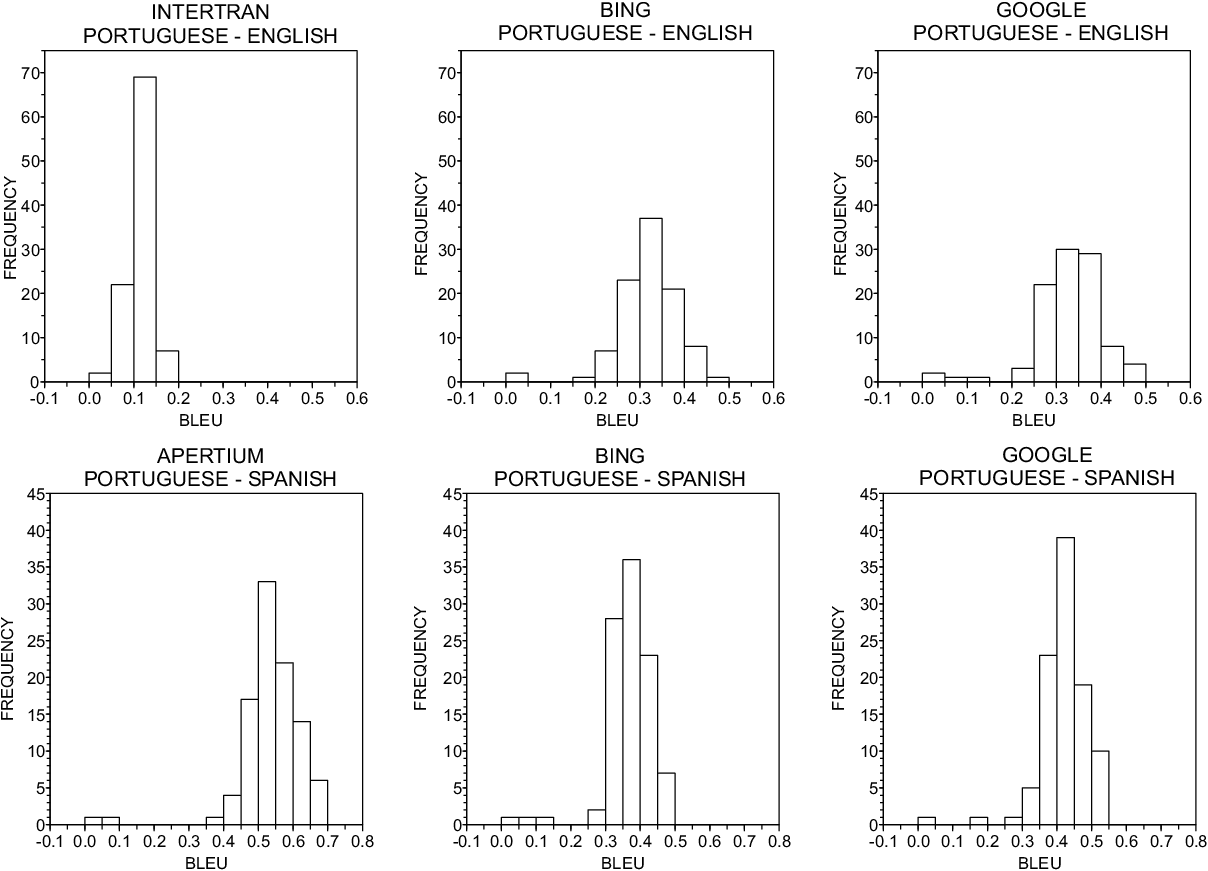}
		\end{center}
		\caption[\it]{\it Distribution of values of BLEU for Portuguese - English (top panel) and Portuguese - Spanish (bottom panel) in the corpus. Notice that while Bing and Google show similar distributions, Apertium and Intertran display quite distinct distributions from Bing and Google.}
		\label{fig2}
\end{figure}

For investigating how topology and semantics can be used to classify translations according to the source translator, we employed four translators: InterTran, Google, Bing and Apertium. The database is the same as the one employed in Section \ref{ev_quality}. To compare the ability to distinguish translations using the proposed and the traditional semantic-based indices, we also computed the accuracy rate when NIST and BLEU were used as attributes of machine learning (ML) algorithms~\cite{bishop}. The ML inductors used were: C4.5~\cite{c45}, Naive-Bayes~\cite{naive}, RIPPER~\cite{ripper} and kNN\footnote{In the k-nearest neighbor algorithm we used $k$ ranging from $k$=1 (kNN-1) to $k$=5 (kNN-5).}~\cite{aha}. In special, those algorithms were used because they have been employed to discriminate texts in previous applications~\cite{t2}.  To evaluate the quality, we used the $10$-fold-cross validation strategy~\cite{cross}, which continuously selects $9$ folds from the training set (the choice is made randomly) to train the inductors and uses the remaining fold to evaluate the classifier generated. To illustrate how difficult it is to distinguish the translators, we analyzed the distribution of BLEU for each translator in Figure \ref{fig2}. The conventional indices provide excellent distinction between Apertium/InterTran and Google/Intertran, but there is some overlap between Google and Bing, for both pairs of languages. Therefore, it is expected that the topological (dis)similarity indices could be useful to enhance the distinction among these translators.

The results concerning the Spanish - Portuguese pair are illustrated in Table \ref{tab.11}, which reveals that the traditional metrics (BLEU and NIST) are outperformed by other metrics including topology. There is a considerable difference between the accuracy rate for the similarity based on the slope (which uses both semantic and topological information) and BLEU. Even the similarity based solely on topology (topological measurements) outperformed the BLEU metric. It appears that the topological analysis is able to provide useful information that was not possible to grasp by the traditional semantic analysis. Interestingly, the metric based on motifs, which is also completely based on topology (the label of the nodes is not employed to detect motifs), reached only 37 \% accuracy at best. This means that probably other motifs will need to be introduced in the analysis if the performance is to be improved.

\begin{table}
	\centering
	\caption{\label{tab.11} Accuracy rate to distinguish machine translations from Google, Apertium and Bing using several similarity indices. The texts were translated from Spanish into Portuguese and the similarity indices were used to quantify the difference between the machine translation and the translation taken as a reference (human translation). Semantic, Topologic and Hybrid measures are identified as (S), (T) and (S+T), respectively.}
		\begin{tabular}{|c|c|c|}
			\hline
			\textbf{Similarity Index}	&\textbf{Accuracy Rate}	&\textbf{ML Algorithm}  \\
			\hline
			Pearson (S)					&	65 \%		&	kNN-5	\\
			Slope (S+T)						&	63	\%		&	Naive	Bayes	\\
			Cosine (S)					&	60 \%		&	Naive Bayes	\\
			Euclidean (S)				&	59 \%		&	Ripper		\\
			Semantic Matching (S)		&	59 \%		&	C4.5			\\
			Topologic Measures (T)	&	58 \%		&	C4.5			\\
			Topologic Matching (S+T)&	53 \%		&	C4.5			\\
			\textbf{BLEU}	(S)		&	\textbf{51 \%}		& 	C4.5	\\
			\textbf{NIST}	(S)		&	\textbf{50 \%}		&	C4.5	\\
			Katz	 (S+T)					&	48 \%		&	kNN-5			\\
			Motifs (T)					&	45 \%		&	Ripper		\\
			Leicht-Holme-Newman	(S)   &	37 \%		&	kNN-1			\\
			\hline
		\end{tabular}
\end{table}

Table \ref{tab.2} summarizes the best accuracy rates in the distinction between Apertium and Google. As expected, the rate increased substantially, since the quality of Google and Apertium are quite different, as shown in Figure \ref{fig2} (the difference can also be easily noticed by manually inspecting short translated extracts). As for the similarity indices, the measure based on the Slope index again achieved the best results, although very close to the BLEU accuracy rate. This result confirms that, even comparing different quality of translations from the semantic point of view, it is still possible to improve the ability to characterize text through the addition of topology-based metrics. With regard to other similarity metrics, their ranking in Table \ref{tab.2} seems to have been maintained when compared with the ranking in Table \ref{tab.11}.

\begin{table}
	\centering
	\caption{\label{tab.2}Accuracy rate to distinguish machine translations from Google and Apertium using several similarity indices. The texts were translated from Spanish into Portuguese and the similarity indices were used to quantify the difference between the machine translation and the translation taken as a reference (human translation). Semantic, Topologic and Hybrid measures are identified as (S), (T) and (S+T), respectively.}
		\begin{tabular}{|c|c|c|}
			\hline
			\textbf{Similarity Index}	&\textbf{Accuracy Rate}	&\textbf{ML Algorithm}  \\
			\hline
			Slope	(S+T)					&	88	\%		&	C4.5	\\
			\textbf{BLEU} (S)			&	\textbf{84 \%}		&	Naive	Bayes \\
			Euclidean (S)				&	82 \%		&	kNN-5	\\
			Cosine	(S)				&	81 \%		&	kNN-5	\\
			Pearson	(S)				&	80 \%		&	C4.5	\\
			Semantic Matching (S)		&	77 \%		&	C4.5	\\
			Topologic Matching (S+T)	&	75 \%		&	Ripper\\
			Topologic Measures (T)	&	74 \%		&	kNN-5	\\
			Katz (S+T)						&	69 \%		&	kNN-3	\\
			\textbf{NIST} (S)			&	\textbf{67 \%}	&	C4.5	\\
			Leicht-Holme-Newman (S)	&	59 \%		&	kNN-3 \\
			Motifs					&	59 \%		&	kNN-5	\\
			\hline
		\end{tabular}
\end{table}

A similar behavior was observed for the translations involving the English language. For example, Table \ref{tab.3}, which illustrates the best accuracy rates in the classification between InterTran, Bing and Google, shows that the accuracy rates of Cosine, Euclidean and Pearson similarities remained quite close. Also, the metric based on motifs still led to the worst accuracy rates. On the other hand, the Katz and semantic matching similarity, which are based on both semantics and topology, displayed the highest accuracy rates, along with the BLEU measure. The same applies to Table \ref{tab.4} in the ability to distinguish between Google and InterTran.

In summary, the experiments with translation confirmed the hypothesis that topological measurements in conjunction with semantic features are able to improve the quantification of similarity in written texts, even if to a small extent. Taken separately, the purely semantic metrics (such as the similarity based on the Pearson coefficient) outperformed the metrics based exclusively on topological features (such as topological measurements and motifs). This confirms that the nature of the problem studied is mainly semantic. In fact, it seems that the main factor in distinguishing quality is the use of correct words, and this is the probable reason why the semantic analysis has become the standard analysis~\cite{eval}. To illustrate a scenario where both structure and semantics can be used for different purposes, we apply in the next section semantic and stylistic indices to detect authorship in poetry and prose.

\begin{table}
	\centering
	\caption{\label{tab.3}Accuracy rate to distinguish machine translations from Google, Intertran and Bing. The texts were translated from Portuguese into English and the similarity indices were used to quantify the difference between the machine translation and the translation taken as a reference (human translation). Semantic, Topologic and Hybrid measures are identified as (S), (T) and (S+T), respectively.}
		\begin{tabular}{|c|c|c|}
			\hline
			\textbf{Similarity Index}	&\textbf{Accuracy Rate}	&\textbf{ML Algorithm}  \\
			\hline
			Semantic Matching (S)		&	68 \%		&	Ripper	\\
			Katz (S+T)						&	67 \%		&	Ripper	\\
			\textbf{BLEU} (S)			&	\textbf{67 \%}	&	kNN-5	\\
			Cosine (S)					&	65 \%		&	Naive Bayes	\\
			Pearson (S)					&	65 \%		&	Naive Bayes	\\
			Euclidean (S)				&	65 \%		&	Naive Bayes	\\
			\textbf{NIST} (S)			&	\textbf{64 \%}	&	Naive Bayes	\\
			Leicht-Holme-Newman (S)						&	62 \%		&	Naive Bayes	\\
			Slope (S+T)						&	53	\%		&	Naive	Bayes	\\
			Topologic Measures (T)	&	51 \%		&	Naive Bayes	\\
			Topologic Matching (S+T)	&	43 \%		&	Naive Bayes	\\
			Motifs	(T)				&	40 \%		&	kNN-5			\\
			\hline
		\end{tabular}
\end{table}

\begin{table}
	\centering
	\caption{\label{tab.4}Accuracy rate to distinguish machine translations from Google and Intertran. The texts were translated from Portuguese into English and the similarity indices were used to quantify the difference between the machine translation and the translation of reference (human translation).}
		\begin{tabular}{|c|c|c|}
			\hline
			\textbf{Similarity Index}	&\textbf{Accuracy Rate}	&\textbf{ML Algorithm}  \\
			\hline
			Cosine (S)					&	96 \%		&	kNN-5	\\
			Pearson (S)				&	96 \%		&	kNN-5	\\
			Euclidean (S)				&	96 \%		&	Ripper\\
			Semantic Matching (S)		&	95 \%		&	Naive Bayes	\\
			Katz	(S+T)					&	95 \%		& 	Naive Bayes	\\
			\textbf{BLEU} (S)			&	\textbf{95 \%}	&	kNN-5	\\
			\textbf{NIST}	(S)		&	\textbf{92 \%}	&	kNN-5	\\
			Leicht-Holme-Newman (S)	&	90 \%		&	Naive Bayes \\	
			Slope	(S+T)				&	80	\%		&	Naive Bayes	\\
			Topologic Measures (T)	&	75 \%		&	kNN-5	\\
			Topologic Matching (S+T)	&	60 \%		&	Naive Bayes\\
			Motifs	(T)				&  59 \%		&	kNN-5	\\
			\hline
		\end{tabular}
\end{table}

\subsection{Topological Similarity and Applications Related to the Text Style}

As a third application, we examined how the structure and semantics are interrelated in the task of recognition of authorship.
Two corpora were used in this experiment, one of poetry and another with prose. Several poems by Emily Dickinson, Alfred Tennyson, Dylan Thomas and Walt Whitman were obtained from an online repository~\cite{famous}. Because they are generally short, in some cases poems by the same author were juxtaposed to obtain sufficiently long texts for the statistical analysis. As for the corpus of texts in the prose format, we collected 5 books from the Gutenberg Project~\cite{gutenberg} for each of the following authors: Arthur Conan Doyle, Charles Darwin, Thomas Hardy and Bram Stoker. More specifically, we used the first $18,000$ words of each book to build the networks and compute (dis)similarity indices.

\begin{figure}[h]
		\begin{center}
			\includegraphics[width=1\textwidth]{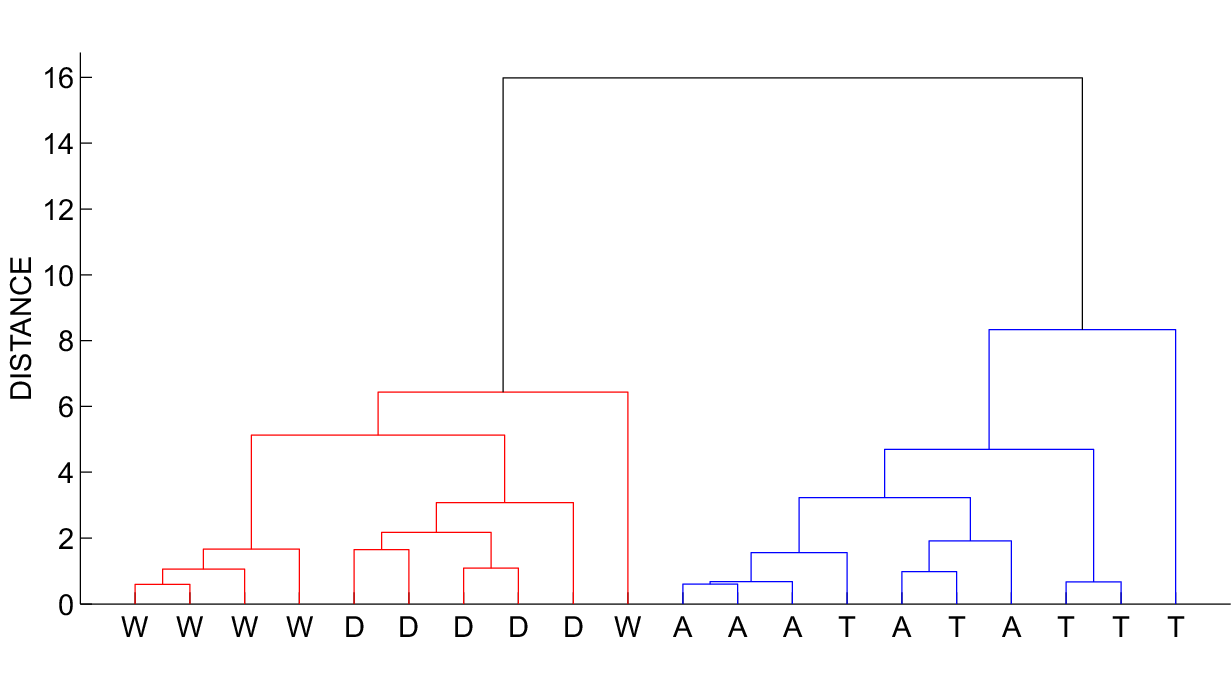}
		\end{center}
		\caption[\it]{\it Hierarchical clustering obtained using structural features to distinguish between Whitman (W), Thomas (T), Tennynson (A) and Dickinson (D).}
		\label{fig4}
\end{figure}

\begin{figure}[h]
		\begin{center}
			\includegraphics[width=1\textwidth]{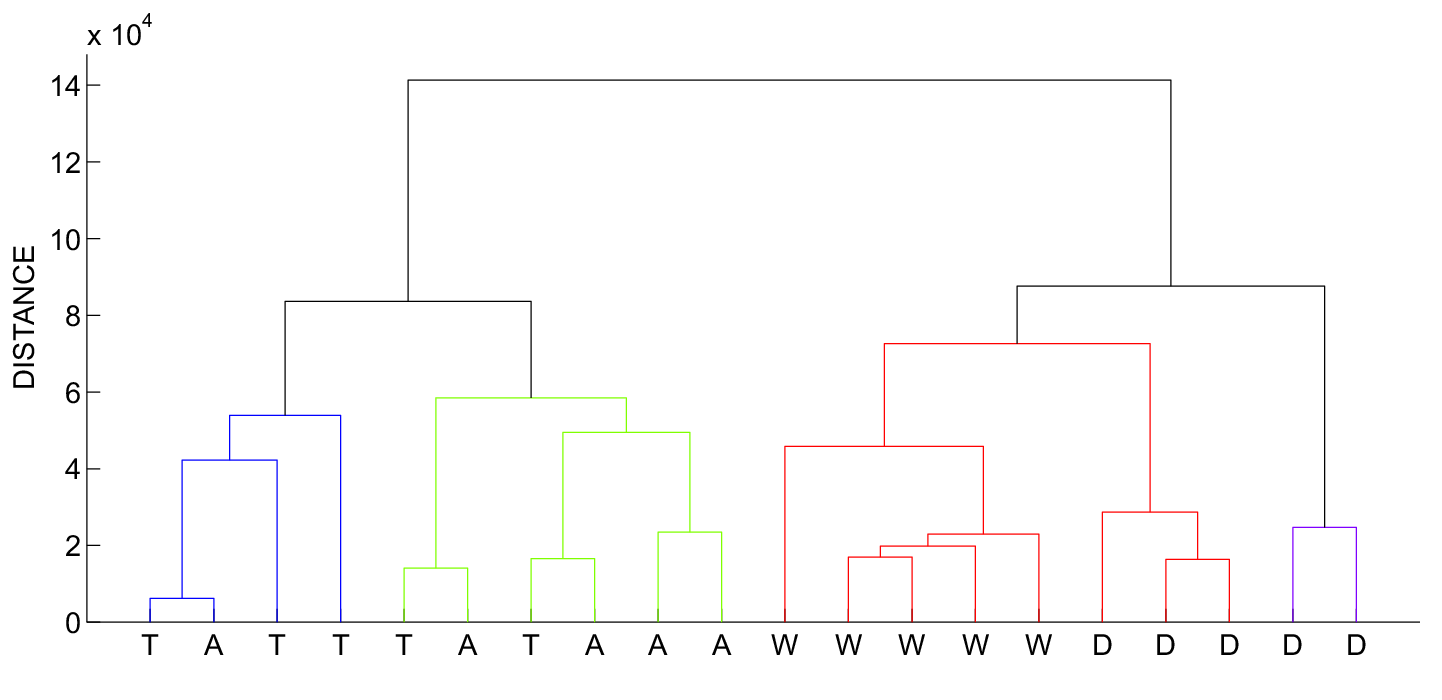}
		\end{center}
		\caption[\it]{\it Hierarchical clustering obtained using semantic features to distinguish between Whitman (W), Thomas (T), Tennynson (A) and Dickinson (D)}
		\label{fig5}
\end{figure}

We applied the dissimilarity metric based on the Euclidean distance (see eq. (\ref{euclideano}) in Section \ref{euclideana}), which is based on semantic features; and the dissimilarity metric based on topological features (see Section \ref{topIndexes}) to recognize authorship in poems. The hierarchy obtained with the Ward method~\cite{ward} using the topological and semantic dissimilarity metrics are illustrated in Figure \ref {fig4} and \ref{fig5}, respectively. The ability to distinguish authors seems to be equivalent. It turns out that for this literary style, both semantic and structure appear to be relevant factors to characterize authorship. Interestingly, these results are consistent with those from the evaluation of machine translation, once semantics and structure are roughly equivalent.

\begin{figure}[h]
		\begin{center}
			\includegraphics[width=1\textwidth]{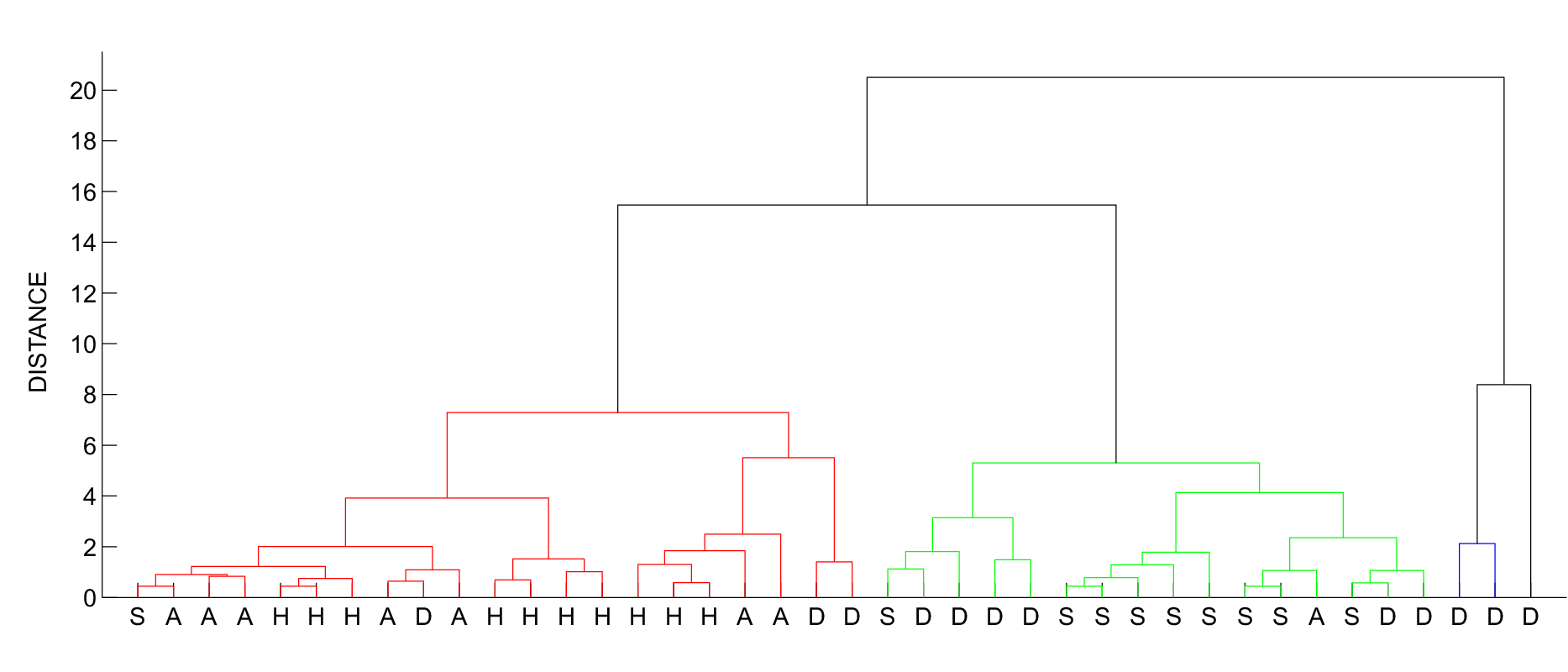}
		\end{center}
		\caption[\it]{\it Hierarchical clustering using structural features. The authors in the hierarchy are: Doyle (A), Darwin (D), Hardy (H) and Stoker (S).}
		\label{fig6}
\end{figure}

\begin{figure}[h]
		\begin{center}
			\includegraphics[width=1\textwidth]{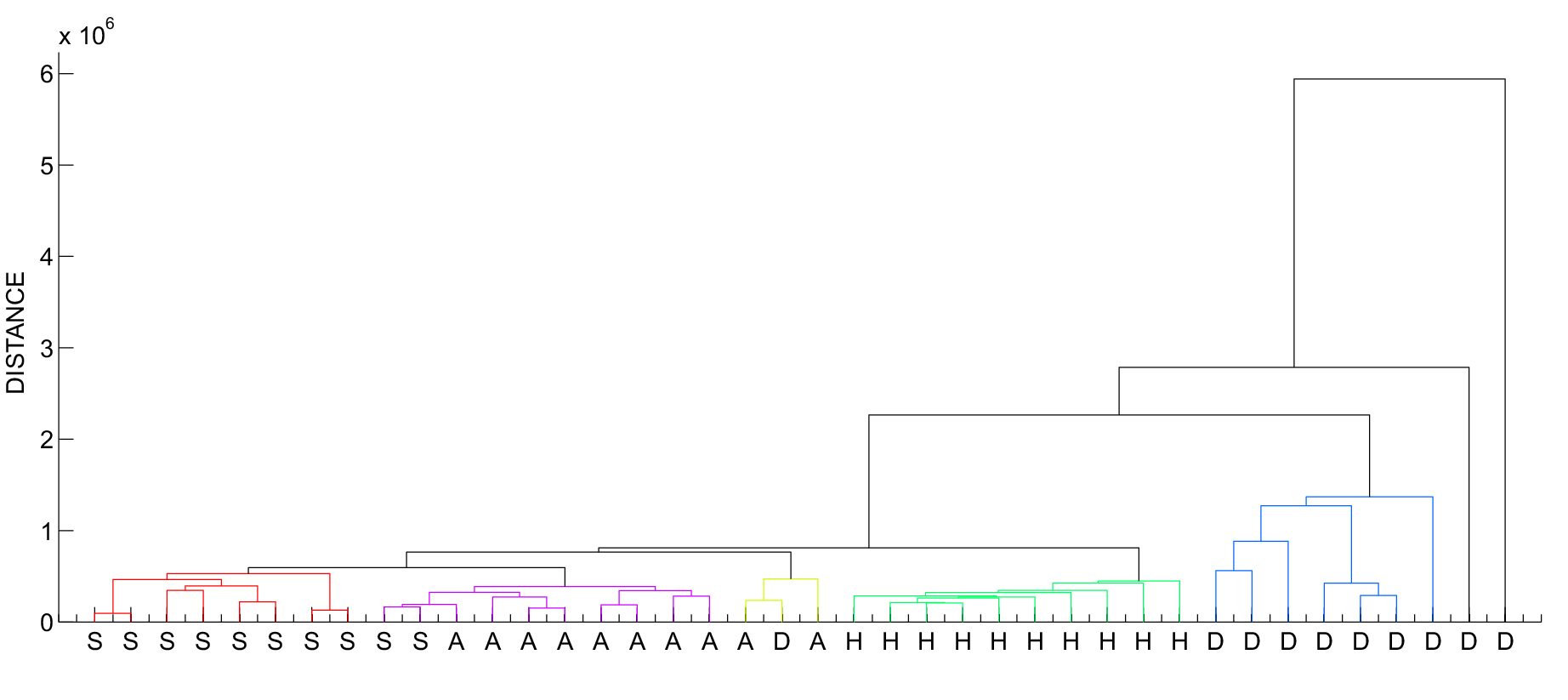}
		\end{center}
		\caption[\it]{\it  Hierarchical clustering using semantic features. The authors in the hierarchy are: Doyle (A), Darwin (D), Hardy (H) and Stoker (S). }
		\label{fig7}
\end{figure}

A second experiment on authorship recognition was carried out with a corpus on prose involving four authors of story books. The hierarchies obtained are illustrated in Figures \ref{fig6} and \ref{fig7}, which point to a high correlation between the semantic and topological paradigms, since the ability to distinguish among authors is quite similar. However, a more refined analysis indicates that different patterns do emerge. Consider, for example, Stoker and Darwin. While the topological analysis reveals that they display similar writing styles, the semantic contents in their texts are quite different. In other words, the writing style is shared, even though they write about completely different subjects~\footnote{Further analyzing the works of both authors, we confirmed that the themes developed by each one are different. For Stocker wrote story books, while Darwin compiled scientific manuscripts.}. Analogously, Stoker and Doylan shared semantic contents, but used different styles. Overall, the suitable methods to use in the classification depend on the purpose, whether one wishes to distinguish writing style or topics. Furthermore, semantic and topological features may be combined to identify authors when many authors are to be distinguished. For authors with the same style can eventually be distinguished by the semantic contents, while authors who write about the same topic can be distinguished by the individual subtleties of style.

\section{Conclusion} \label{conclusion}

In this paper we considered the problem of measuring similarity between pairs of texts, which is relevant in many situations of linguistic interest and have significant consequences for our understanding of textual phenomena. \Red{Many studies have been made for quantifying content similarity and classifying texts, but to the best of our knowledge this paper is the first to combine semantic features and topology of complex networks to enhance the performance of real applications. We performed a systematic evaluation for three natural language processing tasks, namely identification of machine translation systems, evaluation of quality of machine translated texts and authorship recognition.} More specifically, applying the concepts and methodologies of complex networks to characterize texts according to the stylistic features, we proposed and evaluated several similarity and dissimilarity indices, some of which did not involve any kind of semantic information. Overall we showed that semantic contents are still the most important feature to define similarity. Nevertheless, for some applications the use of topological metrics may be beneficial, especially if combined with semantic evaluation. Of particular importance was the finding that the number of paths between concepts, the standard deviation of betweenness and out-degree seem to be good indicators of translation quality. Furthermore, in authorship recognition topological features may be key to distinguishing styles. \Red{In fact, we have found that the relationship between authors' manuscripts mainly depends on the nature of the similarity index as semantically related authors may have developed completely different writing styles and vice-versa.} It is hoped that the approach suggested here may lead to the development of more robust, efficient similarity indices, and boost research of other areas where topology has been shown to effectively characterize written texts.


\section*{Acknowledgments}

This work was supported by FAPESP and CNPq (Brazil).

\end{document}